# Ensemble data assimilation to diagnose AI-based weather prediction model: A case with ClimaX version 0.3.1

Shunji Kotsuki[1,2,3], Kenta Shiraishi[4], and Atsushi Okazaki[1,2]

[1]Institute for Advanced Academic Research, Chiba University, Chiba, Japan
[2]Center for Environmental Remote Sensing, Chiba University, Chiba, Japan
[3]Research Institute of Disaster Medicine, Chiba University, Chiba, Japan
[4]Graduate School of Science and Engineering, Chiba University, Chiba, Japan

*Correspondence to*: Shunji Kotsuki (shunji.kotsuki@chiba-u.jp)

**Abstract.**

Artificial intelligence (AI)-based weather prediction research is growing rapidly and has shown to be competitive with the advanced dynamic numerical weather prediction models. However, research combining AI-based weather prediction models with data assimilation remains limited partially because long-term sequential data assimilation cycles are required to evaluate data assimilation systems. This study proposes using ensemble data assimilation for diagnosing AI-based weather prediction models, and marked the first successful implementation of ensemble Kalman filter with AI-based weather prediction models. Our experiments with an AI-based model ClimaX demonstrated that the ensemble data assimilation cycled stably for the AI-based weather prediction model using covariance inflation and localization techniques within the ensemble Kalman filter. While ClimaX showed some limitations in capturing flow-dependent error covariance compared to dynamical models, the AI-based ensemble forecasts provided reasonable and beneficial error covariance in sparsely observed regions. In addition, ensemble data assimilation revealed that error growth based on ensemble ClimaX predictions was weaker than that of dynamical NWP models, leading to higher inflation factors. A series of experiments demonstrated that ensemble data assimilation can be used to diagnose properties of AI weather prediction models such as physical consistency and accurate error growth representation.

# 1 Introduction

The intensification of weather-induced disasters due to climate change is becoming increasingly severe worldwide (e.g., Jonkman et al. 2024). In a recent risk report, the World Economic Forum (2023) indicated that extreme weather is among the most severe global threats. To address extreme weather events such as torrential heavy rains and heat waves, further advancements in weather forecasting are essential. There are two essential components for accurate weather forecasting: (1) numerical weather prediction (NWP) models that forecast future weather based on initial conditions, and (2) data assimilation, which integrates atmospheric observation data to estimate initial conditions for subsequent forecasts by NWP models.

Since Google DeepMind issued the first artificial intelligence (AI) weather prediction model, GraphCast, in December 2022 (Lam et al. 2023), deep learning-based weather prediction research has shown rapid growth. A number of AI weather prediction models have been proposed even by private information and technology (IT) companies such as Pangu-Weather by Huawei (Bi et al. 2023), ClimaX and Stormer by Microsoft (Nguen et al. 2023, 2024), and FourCastNet by NVIDIA (Pathak et al. 2022, Bonev et al. 2023). These machine learning approaches have been shown to be competitive with state-of-the-art NWP models (e.g., Kochkov et al. 2024). Progresses in AI-based weather prediction has been supported by the expansion of benchmark data and evaluation algorithms, such as WeatherBench (Rasp et al. 2020, 2024). Notably, most AI-based weather prediction models, including Pang-Weather, ClimaX, Stormer, and FourCastNet, use the Vision Transformer (ViT) neural network architecture (Vaswani et al. 2017, Dosovitski et al. 2020). The ViT, which has been explored in language models and image classifications, was demonstrated to be effective in weather prediction as well.

However, research that couples AI-based weather prediction models with data assimilation remains limited. This limitation is partially due to the fact that long-term sequential data assimilation experiments are needed for the evaluation of data assimilation systems, in contrast to weather prediction tasks that allow for parallel learning using benchmark data. Conventional data assimilation methods used in NWP systems can be categorized into three groups: variational methods, ensemble Kalman filters, and particle filters. There are strong mathematical similarities between neural networks and variational data assimilation, both of which minimize their cost functions using their differentiable models. Because auto-differentiation codes are always available for neural-network-based AI models, AI weather prediction models are considered compatible with variational data assimilation methods as in Xiao et al. (2023) and Adrian et al. (2024). On the other hand, recent studies have started to solve the inverse problem inherent in data assimilation by deep neural networks (Chen et al. 2023, Vaughan et al. 2024). However, no study has succeeded in employing ensemble Kalman filtering with AI models. Since AI models require significantly lower computational costs compared to dynamical NWP models, ensemble-based methods, such as ensemble Kalman filters (EnKFs) and particle filters, also offer benefits for AI models.

This study proposes using ensemble data assimilation for diagnosing AI-based weather prediction models. For that purpose, this study marks the first successful implementation of ensemble Kalman filter experiments with an AI weather prediction model to the best of the authors knowledge. We applied the ViT-based ClimaX (Nguen et al. 2023) to data assimilation experiments using the available source code and experimental environments with necessary modifications. For data assimilation, we applied the local ensemble transform Kalman filter (LETKF) (Hunt et al. 2007), which is among the most widely used data assimilation methods in operational NWP centers such as the European Centre for Medium-Range Weather Forecasts (ECMWF), Deutscher Wetterdienst (DWD) and Japan Meteorological Agency (JMA). Using the coupled ClimaX–LETKF data assimilation system, we investigated several key aspects of AI-based weather prediction model, including whether the data assimilation cycles stably for the ClimaX AI weather prediction model using ensemble Kalman filters; whether AI-based ensemble weather prediction accurately represents flow-dependent background error variance and covariance. We also investigated whether techniques such as covariance inflation and localization, which are conventionally used in EnKFs for dynamical NWP models, are effective for AI weather prediction models. By addressing these research questions, we aim to advance the integration of AI weather prediction models with data assimilation techniques, toward the development of more accurate weather forecasting.

The rest of paper is organized as follows: section 2 describes the methods and experiments and section 3 presents the results. Finally, section 4 provides discussion and summary.

## 2 Methods and experiments

### 2.1 ClimaX Model

The ClimaX (Nguen et al. 2023) is a ViT-based AI weather prediction model for the global atmosphere. Variable tokenization and variable aggregation are the key components of the ClimaX architecture upon ViT, as they provide flexibility and generality. This study used the low-resolution version of ClimaX (version 0.3.1), with 64 and 32 zonal and meridional grids, respectively, corresponding to a spatial resolution of 5.625° × 5.625°. The vertical model level was set at seven (900, 850, 700, 600, 500, 250 and 50 hPa).

By default, ClimaX is set to be trained against only five variables: geopotential at 500 hPa, temperature at 850 hPa, temperature at 2 m, zonal wind at 10 m, and meridional wind at 10 m. We updated ClimaX for data assimilation, which allowed the AI model to produce variables required for subsequent forecasts (Table 1). We also diagnosed surface pressure, which is a required input for data assimilation, based on geopotential and surface elevation. Figure 1 shows the training curves of the default and updated ClimaX models verified against WeatherBench data (Rasp et al. 2020). Data for the period 2006–2015 were used for training, and data for 2016 were used for validation. Anomaly correlation coefficients increased and root mean square errors (RMSEs) decreased in Figure 1, indicating successful training of the updated ClimaX model. Because more variables were predicted by the updated ClimaX than by the default ClimaX, more training steps were required.

### 2.2 Local Ensemble Transform Kalman Filter (LETKF)

The LETKF is among the most widely used data assimilation methods in operational NWP centers such as ECMWF, DWD and JMA. The LETKF simultaneously computes analysis equations at every model grid point with the assimilation of surrounding observations within the localization cut-off radius. The ClimaX–LETKF system was developed based on the SPEEDY–LETKF system (Kotsuki et al. 2022) by replacing the SPEEDY weather prediction model with ClimaX. Our future research can readily be expanded to particle filter experiments because the Kotsuki et al. (2022) system includes local particle filters in addition to the LETKF.

Let $\mathbf{X}_t \equiv \{\mathbf{x}_t^{(1)}, \ldots, \mathbf{x}_t^{(m)}\}$ be an ensemble state matrix, whose ensemble mean and perturbation is given by $\bar{\mathbf{x}}_t$ ($\in \mathbb{R}^n$) and $\delta\mathbf{X}_t \equiv \{\mathbf{x}_t^{(1)} - \bar{\mathbf{x}}_t, \ldots, \mathbf{x}_t^{(m)} - \bar{\mathbf{x}}_t\}$ ($\in \mathbb{R}^{n \times m}$), respectively. Here, $n$ and $m$ are the system and ensemble sizes. The superscript ($i$) and subscript $t$ denote the $i$th ensemble member and indicates the time, respectively. The EnKFs, including LETKF, estimate error covariance $\mathbf{P}$ ($\in \mathbb{R}^{n \times n}$) according to sample estimates based on ensemble perturbation:

$$\mathbf{P} \approx \frac{1}{m-1} \delta\mathbf{X}\delta\mathbf{X}^T. \tag{1}$$

The analysis update equation of the LETKF is given by:

$$\mathbf{X}_t^a = \bar{\mathbf{x}}_t^b \cdot \mathbf{1} + \delta\mathbf{X}_t^b \widetilde{\mathbf{P}}_t^a (\mathbf{Y}_t^b)^T \mathbf{R}_t^{-1} \left(\mathbf{y}_t^o - \overline{H_t(\mathbf{X}_t^b)}\right) \cdot \mathbf{1} + \left[(m-1)\widetilde{\mathbf{P}}_t^a\right]^{1/2}, \tag{2}$$

$$\widetilde{\mathbf{P}}_t^a = \left[\frac{(m-1)}{\beta}\mathbf{I} + (\mathbf{Y}_t^b)^T \mathbf{R}_t^{-1} \mathbf{Y}_t^b\right]^{-1}, \tag{3}$$

where, $\widetilde{\mathbf{P}}$ is the error covariance matrix in the ensemble space ($\in \mathbb{R}^{m \times m}$), $\mathbf{Y} \equiv \mathbf{H}\delta\mathbf{X}$ is the ensemble perturbation matrix in the observation space ($\in \mathbb{R}^{p \times m}$), $\mathbf{R}$ is the observation error covariance matrix ($\in \mathbb{R}^{p \times p}$), $\mathbf{y}$ is the observation vector ($\in \mathbb{R}^p$), $H$ is the observation operator that may be nonlinear, $\mathbf{H}$ ($\in \mathbb{R}^{p \times n}$) is the Jacobian of linear observation operator matrix, and $\mathbf{1}$ is a row vector whose all elements are 1 ($\in \mathbb{R}^m$). Here, $p$ is the number of observations. The superscripts $o$, $b$, and $a$ denote the observation, background, and analysis, respectively. The scalar $\beta$ is a multiplicative inflation factor which inflates the background error covariance such that $\mathbf{P}_t^b \to (1+\beta)\mathbf{P}_t^b$. This study uses the Miyoshi (2011)'s approach, which estimates spatially varying inflation factors adaptively based on observation-space statistics (Desroziers et al. 2005).

Localization is a practically important technique for EnKFs to eliminate long-range erroneous correlations due to the sample estimates of $\mathbf{P}$ with a limited ensemble size (Houtekamer and Zhang, 2016). Although a larger localization can spread observation data information for grid points distant from observations, a larger localization scale can yield suboptimal error covariance because of sampling errors. The LETKF inflates the observation error variance to realizes the localization (Hunt et al. 2007) whose function is given by:

$$l = \begin{cases} \exp\left[-\frac{1}{2}\{(d_h/L_h)^2 + (d_v/L_v)^2\}\right] & \text{if } d_h < 2\sqrt{10/3}L_h \text{ and } d_v < 2\sqrt{10/3}L_v \\ 0 & \text{else} \end{cases}, \tag{4}$$

where $l$ is the localization function, and its inverse $l^{-1}$ is multiplied to inflate $\mathbf{R}$ for the localization. Horizontal and vertical distances (km and log(Pa)) from analysis grid point to the observation are defined by $d_h$ and $d_v$ where subscripts $h$ and $v$ denote horizontal and vertical, respectively. Here, $L_h$ and $L_v$ are tunable horizontal and vertical localization scales (km and log(Pa)). The vertical localization scale $L_v$ was set at 1.0 (log Pa) following the method of Kotsuki et al. (2022). Sensitivity to the horizontal localization scale for $L_h$ = 400, 500, 600, 700, and 800 km is investigated in subsequent experiments.

## 2.3 Data assimilation experiments

In this study, all experiments were conducted as simulation experiments by generating observation data from WeatherBench with additions of Gaussian random noises. Although the real observation data was not directly assimilated, the assimilated observations reflect the real atmosphere in this study, in contrast to observing system simulation experiments. To approximate real-world scenarios, we considered radiosonde-like observations to generate atmospheric observation profiles for observing stations (Figure 2). At observing stations, temperature, and zonal and meridional winds were observed at all seven layers, whereas specific humidity was observed at the first to fourth layers. Table 1 shows the standard deviations of the observation errors. The network of observing stations and observation error standard deviations were consistent with those of the SPEEDY–LETKF experiments (Kotsuki et al. 2022, Kotsuki and Bishop 2022). Observation data were produced at 6-h intervals, such that the data assimilation interval was also 6 h.

We employed a series of data assimilation experiments over a year of 2017, which is not used for training and validation of the ClimaX. The ensemble size is 20, and their initial conditions were taken from WeatherBench data in 2006. Data assimilation experimental results were verified against WeatherBench data.

## 3 Results

Figure 3 presents the time series of global-mean root mean square errors (RMSEs) for temperature and geopotential height at the fifth model level, with four different horizontal localization scales ($L_h$). After the initiation of data assimilation, all experiments showed reductions in analysis errors. Experiments with $L_h$ = 500, 600 and 700 km showed stable performance over a period of one year, until the end of 2017. Notably, data assimilation improved not only the observed variable, temperature, but also the unobserved variables, such as geopotential height. This indicates that observation information was propagated to unobserved variables through the data assimilation cycle. In contrast, the experiment with $L_h$ = 800 km exhibited filter divergence after September 2017 due to erroneous error covariance associated with the larger localization scale. In addition, the experiment with $L_h$ = 400 km kept reducing the RMSEs over a year, indicating that a too small localization scale is suboptimal. This implies that ensemble-based error covariance is beneficial to some extent to propagating the impacts of assimilated observation for distant grid points.

Figure 4 shows the global mean RMSEs for zonal wind, meridional wind, temperature, specific humidity, geopotential height, and surface pressure, as a function of the horizontal localization scales averaged over July–December, 2017. At smaller localization scales ($L_h$ = 400 and 500 km), the analysis RMSEs tended to be lower than the first-guess RMSEs, which suggests that data assimilation was beneficial in reducing errors. Conversely, at larger localization scales ($L_h$ = 700 and 800 km), analysis RMSEs tended to be higher than the first guess RMSEs, indicating that data assimilation degraded the analysis, presumably also due to excessive error covariance at larger localization scales. Among the five experiments, a localization scale of $L_h$ = 600 km yielded the lowest analysis RMSEs for most variables. Significant analysis error reductions were observed for temperature and surface pressure. However, no clear impacts were observed for zonal and meridional winds. Even slight degradations were detected, implying that spatial and inter-variable error covariance may not be well represented in our ClimaX-LETKF.

Here, we investigate the spatial patterns of the difference between the analysis and first-guess mean absolute errors, which is given by:

$$MAE_{diff} = \frac{1}{N_t}\sum_t |\bar{\mathbf{x}}_t^a - \mathbf{x}_t^{WB}| - |\bar{\mathbf{x}}_t^b - \mathbf{x}_t^{WB}|, \quad (5)$$

where $N_t$ is the sample size and superscript WB represents WeatherBench data. Negative and positive values indicate improvements and degradations due to data assimilation. Figure 5 shows the $MAE_{diff}$ for four variables (zonal wind at 850 hPa, temperature at 700 hPa, geopotential height at 500 hPa, and surface pressure) based on the experiments with the localization scale $L_h$ = 500 km, which resulted in RMSE reductions by data assimilation for most of variables in Figure 4. General improvements are seen in grids with observations for zonal wind and temperature (Figs. 5 a and b). However, there were also slight degradations at grids surrounding observing stations, such as those in arctic ocean and along the US and Japanese coasts. We also see degradations for geopotential height in grids where temperature and zonal wind degradations are presented (Fig. 5 c). These degradations suggest ensemble-based spatial error covariance were suboptimal in these regions. In contrast, geopotential height and surface pressure generally improved in the Southern Hemisphere (Figs. 5 c and d). In particular, improvements are seen even at grids surrounding observing stations in the Southern Hemisphere. Specifically, using the spatial and inter-variable error covariance based on AI-based ensemble forecasts was advantageous for geopotential heights and surface pressure in sparsely observed regions.

Another important property is that the ClimaX is less chaotic than dynamical NWP models, as indicated by the estimated inflation factor $\beta$ diagnosed by observation-space statistics (Figure 6). Compared to our study, Kotsuki et al. (2017) estimated much smaller inflation factor for a global ensemble data assimilation system using a dynamical model (cf. Fig. 10a in Kotsuki et al. 2017). The estimated inflation factor of Kotsuki et al. (2017) was substantially smaller than our study. Selz and Craig (2022) noted that an AI-based weather prediction model failed to reproduce rapid initial error growth rates, which would prevent it from replicating the butterfly effect as accurately as dynamical NWP models.

## 4 Discussion and summary

The optimal localization scale was very small unexpectedly in Figure 4. Kondo and Miyoshi (2016) pointed out that a larger localization scale is beneficial for low-resolution models and larger ensemble sizes (cf. Table 1 in Kondo and Miyoshi 2016). Our optimal localization scale for the 20-member ClimaX-LETKF was 600 km, which is significantly shorter than the 900-km scale of the 20-member LETKF experiment coupled with a dynamical NWP model (also known as SPEEDY; Molteni 2003) (see Figure 2b in Kotsuki and Bishop 2022). The SPEEDY model resolution (96 × 48 horizontal grids) is finer than that of the ClimaX in this study (64 × 32 horizontal grids), which suggests that ClimaX captures flow-dependent error covariance less effectively than dynamical NWP models. Bonavita (2024) investigated physical realism of the present AI models (FourCastNet, Pangu-Weather and GraphCast), and concluded that AI models are not able to properly reproduce sub-synoptic and mesoscale weather phenomena. The suboptimal flow-dependent error covariance in this study can be attributed to physical inconsistent atmospheric fields of the ClimaX predictions.

It should be noted that we were unable to conduct observation system simulation experiments (k.a. OSSEs), which requires a natural run by ClimaX. This is because ClimaX could not produce long-term forecasts within our experimental configurations. A typical example is shown in Figure 7. The forecasted temperature fields of ClimaX eventually began to deviate from the WeatherBench data with the continuation of 6-h forecasts. Ultimately, ClimaX produced meteorologically unrealistic weather fields, as demonstrated by the very low temperatures in the Pacific Ocean. Because AIs cannot learn physical laws in the absence of specific treatments, they are more likely to produce unrealistic weather fields under previously unencountered weather conditions. In other words, this suggests that ClimaX is unable to return to a meteorologically plausible attractor (or trajectory) while data assimilation enables the ClimaX to synchronize with the real atmosphere. Applying neural networks that are informed or constrained by physical laws would be necessary to conduct observation system simulation experiments for AI-based weather prediction models.

Two major advancements are required for AI-based weather prediction models to improve ensemble data assimilation. First, it is imperative that AI models generate physically consistent forecast variables. The accuracy of spatial and inter-variable error covariance would be improved by this enhancement, which would require AI model training procedures to include physical constraints such as the hydrostatic and geostrophic balances, in addition to decreasing the mean square errors of the target variables. Second, it is crucial to accurately capture error growth rate. Our findings demonstrated that error growth based on ensemble ClimaX predictions were weaker than those of dynamical NWP models, leading to higher inflation factors (Figure 6). Thus, ensemble forecasts produced by AI weather prediction models likely exhibit insufficient spread. In weather forecasting, capturing forecast uncertainty is as important as providing accurate forecasts. One possible solution for improving the error growth is to develop a set of slightly different AI models by randomizing the seed in the AI training process as an analog of stochastic parameterization (Weyn et al. 2021).

Despite the need for further improvements, this study represents a significant step toward ensemble data assimilation for AI-based weather prediction models. Notably, we demonstrated that the data assimilation cycled stably for the AI-based weather prediction model ClimaX with the LETKF using covariance inflation and localization techniques. In addition, the ensemble-based error covariance was reasonable in sparsely observed regions, even according to AI weather prediction models.

Additional research is anticipated for areas identified as requiring further improvements. For that purpose, ensemble data assimilation is a useful tool for diagnosing AI-based weather forecasting models. Namely, investigating optimal localization scales, ensemble-based error covariance and necessary inflation factors give beneficial insights to understand properties of AI models. Since AI models require much lower computational costs compared to dynamical NWP models, extending the present study to large-ensemble EnKFs or LPF is also important subjects of future studies.

**Code and data availability**

The data assimilation system, experimental data, and visualization scripts used in this manuscript are archived on Zonodo (https://zenodo.org/records/13884167; doi: 10.5281/zenodo.13884167). The original ClimaX version 0.3.1 and LETKF codes are also archived on Zenodo; ClimaX version 0.3.1 (https://zenodo.org/records/14258100, doi: 10.5281/zenodo.14258099) and LETKF (https://zenodo.org/records/14258014, doi: 10.5281/zenodo.14258014).

**Author contributions**

S. Kotsuki developed the ClimaX-LETKF system and employed data assimilation experiments, K. Shiraishi updated and trained the ClimaX, and A. Okazaki made large amount of discussion about the analyses of the experiments.

**Competing interests**

The authors have no competing interests to declare.

**Acknowledgements**

This study was partly supported by the JST Moonshot R&D (JPMJMS2389), the Japan Society for the Promotion of Science (JSPS) KAKENHI grants JP21H04571, JP21H05002, JP22K18821, and the IAAR Research Support Program and VL Program of Chiba University.

**Tables**

**Table 1**: Variables of the ClimaX model used in this study. ClimaX requires input variables to predict output variables. Observation (Obs) variables are assimilated with associated error standard deviation (Error SD).

| Symbol | Variable | Unit | Input | Output | Obs | Error SD | Height |
|---|---|---|---|---|---|---|---|
| U | Zonal wind | m/s | X | X | X | 1.0 | 925, 850, 700, 600, 500, 250, 50 (hPa) |
| V | Meridional wind | m/s | X | X | X | 1.0 | |
| T | Temperature | K | X | X | X | 1.0 | |
| Q | Specific humidity | kg/kg | X | X | X (※) | 0.1 | |
| Geo | Geopotential | m2/s2 | X | X | | | |
| U10m | 10-m zonal wind | m/s | X | X | | | 10 m |
| V10m | 10-m meridional wind | m/s | X | X | | | 10 m |
| T2m | 2-m temperature | m/s | X | X | | | 2 m |
| Ps | Surface pressure | hPa | | | X | 1.0 | Surface |
| Elev | Surface elevation | m | X | | | | Surface |
| Lon | Longitude | degree | X | | | | – |
| Lat | Latitude | degree | X | | | | – |
| Mask | Land–sea mask | 1 or 0 | X | | | | – |

※ Specific humidity is observed up to 4th model level (i.e., 925, 850, 700 and 600 hPa).

**Figures**

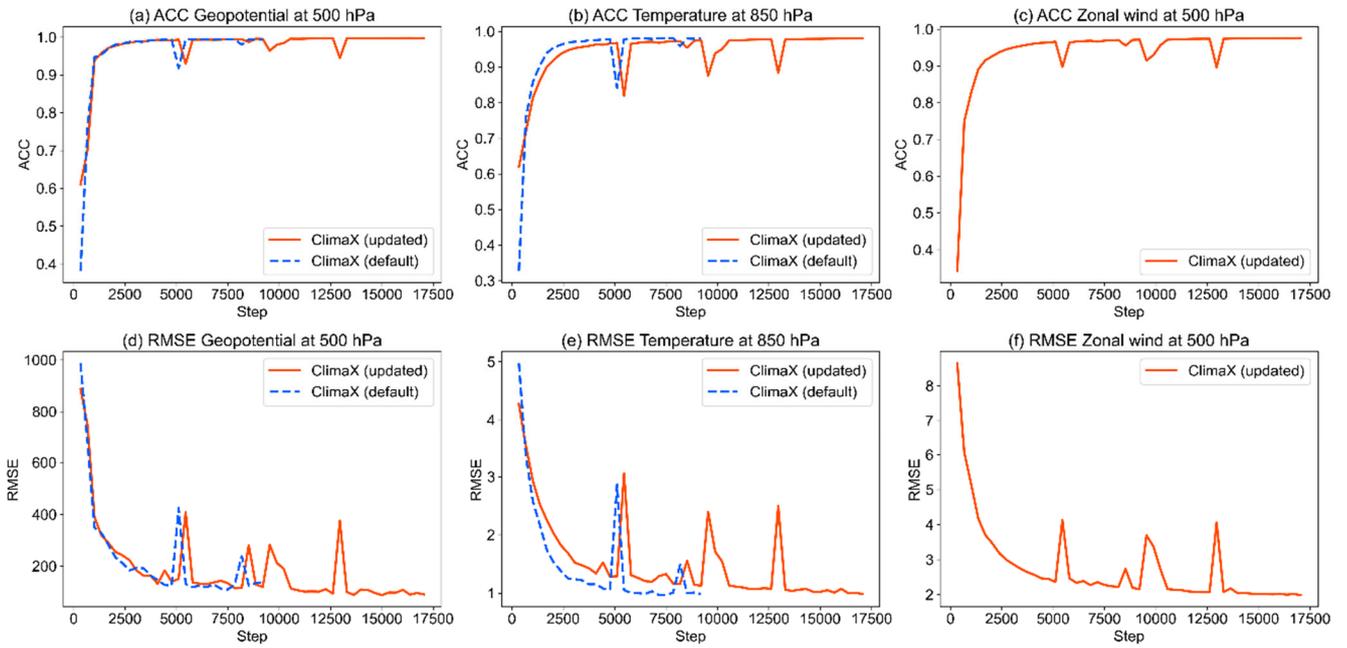

**Figure 1:** Training curves for the default and updated ClimaX models (dashed blue and solid orange lines) verified against WeatherBench data in 2016, as a function of the number of training steps. Each training step includes 64 training data in a mini batch. Panels (a-c) and (d-f) show anomaly correlation coefficients (ACCs) and root mean square errors (RMSEs). (a, d), (b, e) and (c, f) are geopotential at 500 hPa (m2/s2), temperature at 850 hPa and zonal wind at 500 hPa. There are no blue dashed lines in panels (c) and (f) because the default ClimaX model does not predict zonal wind at 500 hPa.

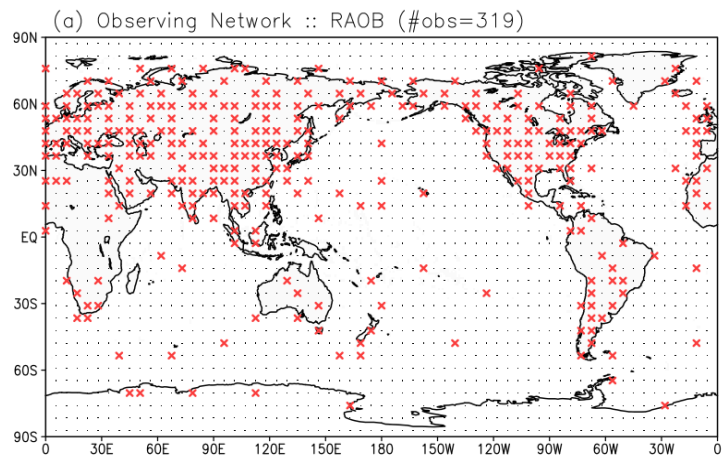

**Figure 2.** The observing network. Small black dots and red crosses represent model grid points and observing points, respectively.

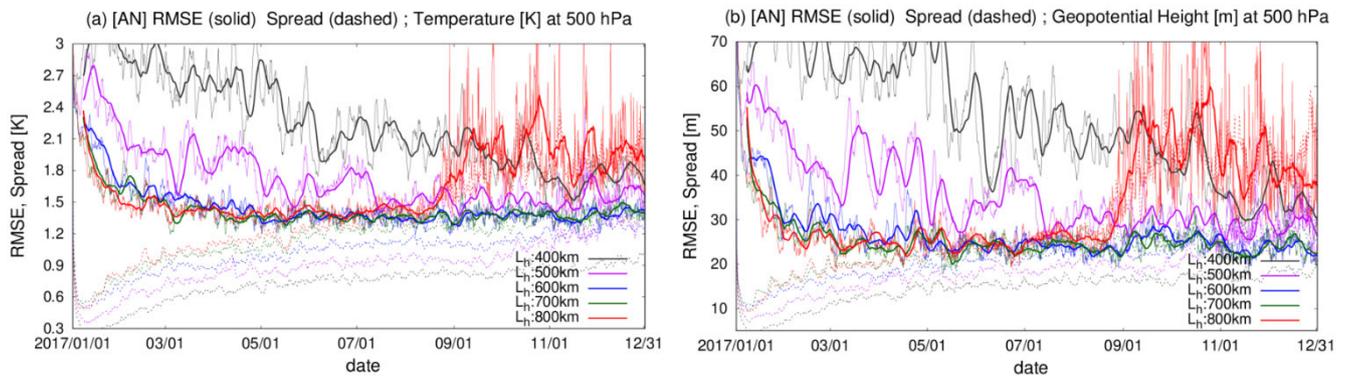

**Figure 3:** Time series of global-mean root mean square errors (RMSEs) verified against WeatharBench data, and ensemble spreads for (a) temperature (K) and geopotential height (m) at the fifth model level (= 500 hPa). Thin and bold solid lines indicate 6-hourly RMSEs and their 7-day running means, respectively. Dashed lines indicate ensemble spreads. Black, purple, blue, green, and red lines indicate the ClimaX-LETKF experiments, at localization scales of $L_h$= 400, 500, 600, 700 and 800 km. The abscissa indicates the date (month/day) in 2017.

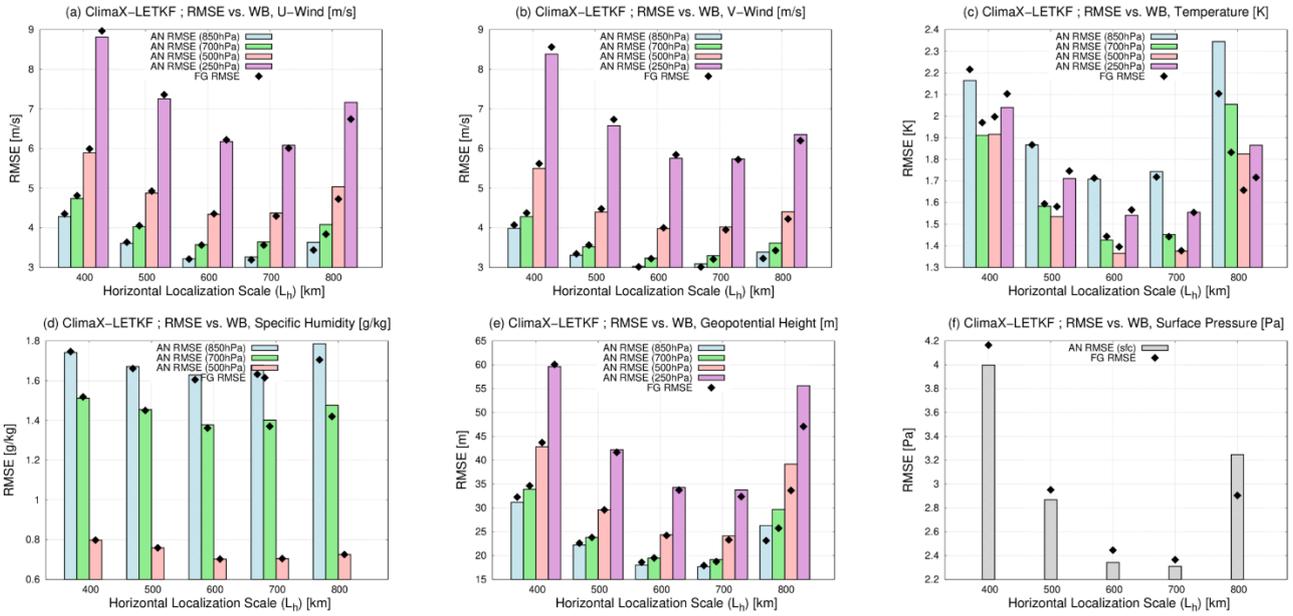

**Figure 4:** Global mean root mean square errors (RMSEs) for (a) zonal wind (m/s), (b) meridional wind (m/s), (c) temperature (K), (d) specific humidity (g/kg), (e) geopotential height (m), and (f) surface pressure (hPa), as a function of the horizontal localization scales (km) averaged over July–December 2017. Colored bars and black diamonds indicate analysis (AN) and first-guess (FG) RMSEs, respectively. Blue, green, red, and purple bars in (a-e) represent 2nd, 3rd, 5th and 6th model levels (850, 700, 500, and 250 hPa, respectively). Gray bars in (f) represent surface pressure. The RMSEs of specific humidity at the 6th model level in (d) were too low to be shown.

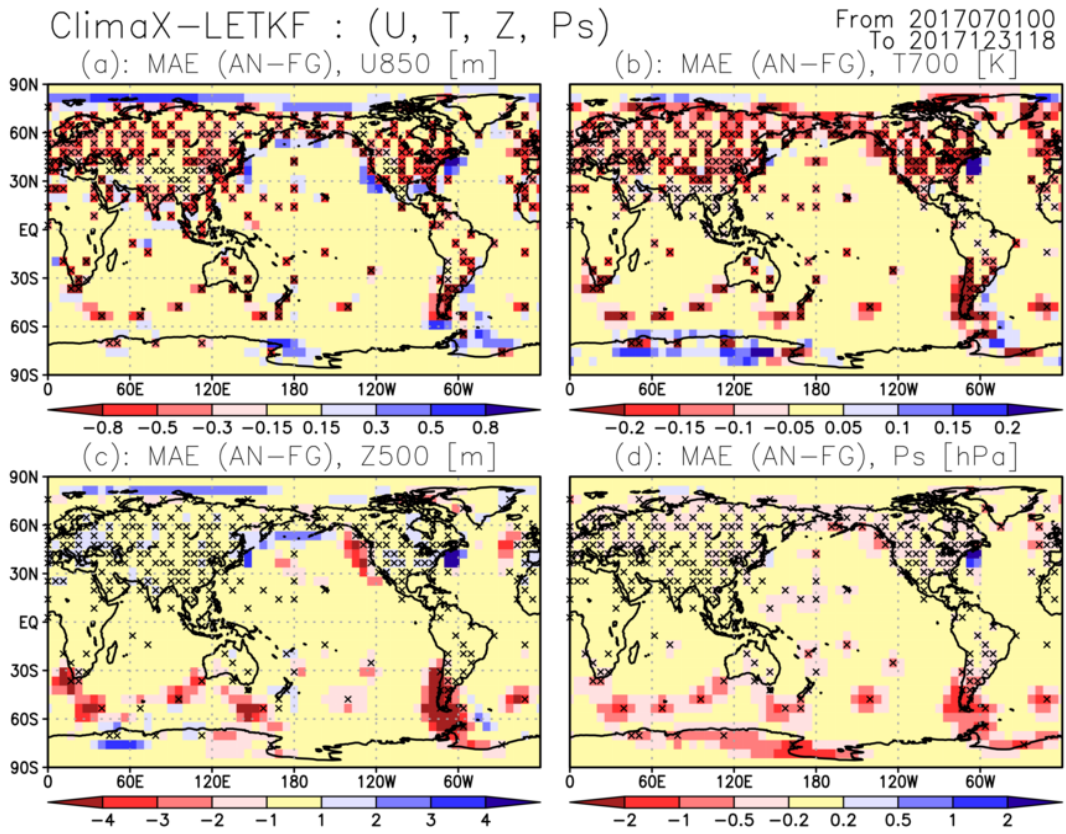

**Figure 5:** Spatial patterns of difference between analysis (AN) and first-guess (FG) mean absolute errors (MAEs) for (a) zonal wind (m/s) at 850 hPa, (b) temperature (K) at 700 hPa, (c) geopotential height (m) at 500 hPa, and surface pressure (hPa), averaged over July–December 2017. Warm and cold colors represent improvements and degradations due to data assimilation. Results are for a localization scale of $L_h$ = 500 km. Black crosses indicate observing stations.

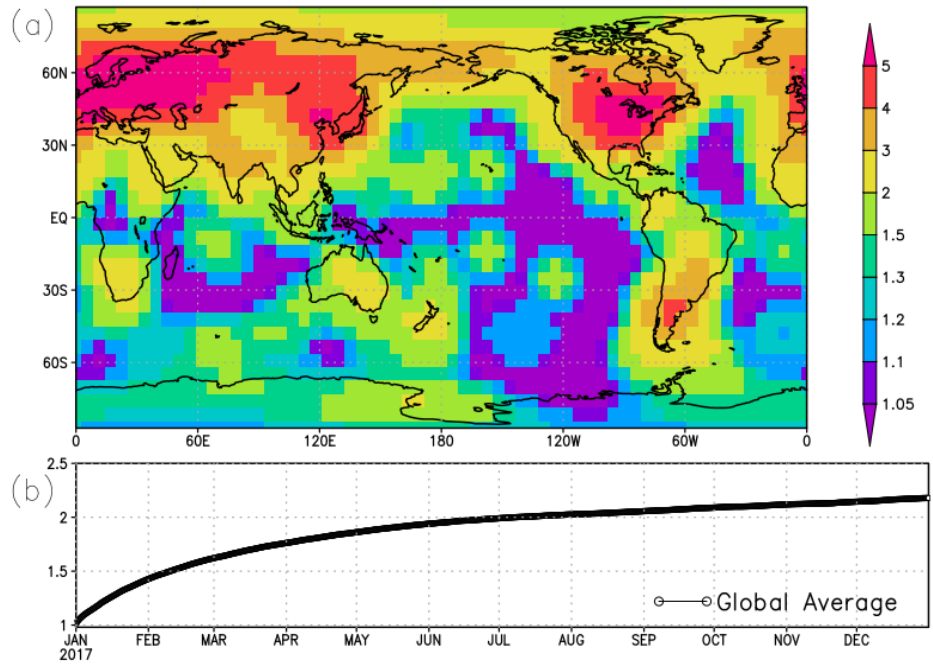

**Figure 6:** (a) Spatial pattern of the multiplicative inflation factor at the end of experiment on 1800UTC of December 31, 2017. (b) Time series of globally averaged inflation factors. Results are for a localization scale of $L_h$ = 600 km.

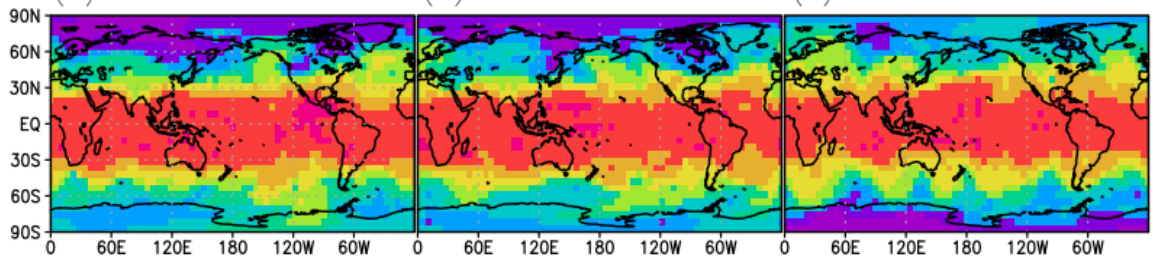

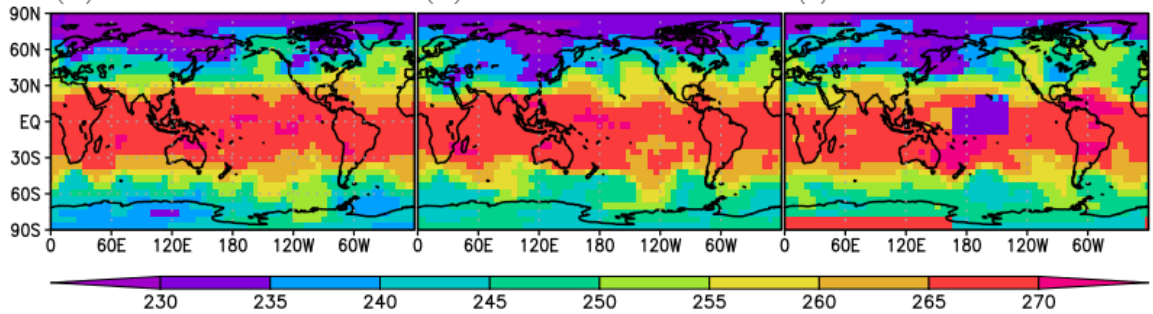

**Figure 7:** Spatial patterns of temperature (K) at 5th model level (500 hPa). Panels (a-c) are WeatherBench data. Panels (d-f) are forecasts by ClimaX initialized at 0000 UTC of January 1, 2017. Panels (a, d) show 0000 UTC of January 3, 2017, (b, e) show 0000 UTC of February 1, 2017, and (c, f) show 0000 UTC of May 1, 2017, respectively.